\documentclass[letterpaper, 10 pt, conference]{ieeeconf}  % Comment this line out if you need a4paper

\IEEEoverridecommandlockouts                              % This command is only needed if 
\usepackage[ruled,linesnumbered,vlined]{algorithm2e}

\usepackage{graphics} % for pdf, bitmapped graphics files
\usepackage{epsfig} % for postscript graphics files
\usepackage{times} % assumes new font selection scheme installed
\usepackage{mathrsfs} %引入该宏包
\usepackage{amsmath,amssymb}
\usepackage{circledsteps}
\usepackage{enumerate}
\usepackage{cite}
\usepackage{booktabs}
\usepackage{multirow}
\usepackage{array}
\usepackage{makecell}
\usepackage{threeparttable}
\title{\LARGE \bf

Sparse-Graph-Enabled Formation Planning

for Large-Scale Aerial Swarms
}

\author{Yuan Zhou, Lun Quan, Chao Xu, Guangtong Xu, and Fei Gao% <-this % stops a space
\thanks{This work was supported by the National Natural Science Foundation of China under Grant 62322314 and Grant 62203256. \emph{(Corresponding Author: Guangtong Xu; Fei Gao.)}}
\thanks{Yuan Zhou, Lun Quan, Chao Xu, and Fei Gao are with the Institute of Cyber-Systems and Control, College of Control Science and Engineering, Zhejiang University, Hangzhou 310027, China, and also with the Huzhou Institute, Zhejiang University, Huzhou 313000, China. (e-mail: \{y2zhou, lunquan, cxu, fgaoaa\}@zju.edu.cn). Guangtong Xu is with the Huzhou Institute, Zhejiang University, Huzhou 313000, China. (e-mail: guangtong\_xu@163.com).}
}

\begin{document}

\maketitle
\thispagestyle{empty}
\pagestyle{empty}

%%%%%%%%%%%%%%%%%%%%%%%%%%%%%%%%%%%%%%%%%%%%%%%%%%%%%%%%%%%%%%%%%%%%%%%%%%%%%%%%

\begin{abstract}

The formation trajectory planning using complete graphs to model collaborative constraints becomes computationally intractable as the number of drones increases due to the curse of dimensionality. To tackle this issue, this paper presents a sparse graph construction method for formation planning to realize better efficiency-performance trade-off. Firstly, a sparsification mechanism for complete graphs is designed to ensure the global rigidity of sparsified graphs, which is a necessary condition for uniquely corresponding to a geometric shape. Secondly, a good sparse graph is constructed to preserve the main structural feature of complete graphs sufficiently. 
Since the graph-based formation constraint is described by Laplacian matrix, the sparse graph construction problem is equivalent to submatrix selection, which has combinatorial time complexity and needs a scoring metric. Via comparative simulations, the Max-Trace matrix-revealing metric shows the promising performance.
The sparse graph is integrated into the formation planning. Simulation results with 72 drones in complex environments demonstrate that when preserving 30\% connection edges, our method has comparative formation error and recovery performance w.r.t. complete graphs. Meanwhile, the planning efficiency is improved by approximate an order of magnitude. Benchmark comparisons and ablation studies are conducted to fully validate the merits of our method.

\end{abstract}

%%%%%%%%%%%%%%%%%%%%%%%%%%%%%%%%%%%%%%%%%%%%%%%%%%%%%%%%%%%%%%%%%%%%%%%%%%%%%%%%
\section{Introduction}

Graph-based formation trajectory planning can deal with obstacle avoidance and formation keeping while ensuring flexible formation maneuvers \cite{c1}. To uniquely determine a formation configuration for swarms, the graph describing the connection relationship should be globally rigid~\cite{c2, c3}. In our previous work \cite{c1}, a complete graph is used to guarantee the global rigidity conveniently, and the corresponding Laplacian matrix is adopted to formulate formation constraints. However, since each drone requires considering the trajectories of all others under complete graphs, the computational complexity of formation planning grows quadratically with the increase of swarm scales, inducing prohibitively high computation time for real-time applications.
\begin{figure}[!t]
    \centering
    \includegraphics[width=3.3in]{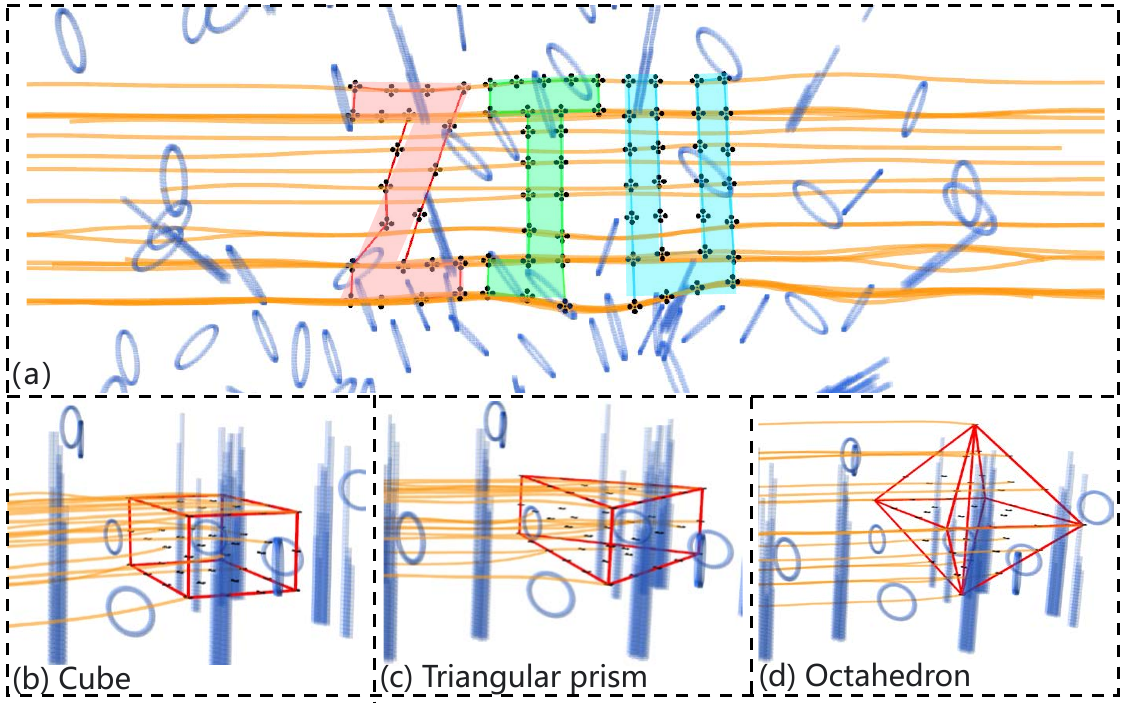}
    \vspace{-0.1cm}
    \caption{Simulation results of the sparse-graph-enabled formation planning.
    (a) The visualization of spelling “ZJU” with 72 drones while avoiding obstacles.
    (b)-(d) The illustration of keeping different geometric shapes with 48 drones in cluttered environments.
    % Please watch our supplemental videos for more information at......
    }
    \vspace{-0.3cm}
    \label{figurelabel}
\end{figure}

Intuitively, if there exists an efficient approach to establish a sparse graph connecting a few edges that captures the main feature of complete graphs, the problem dimension of formation planing will be reduced significantly while preserving formation performance. In graph theory, a minimum number of edges can be found to ensure the global rigidity of graphs theoretically \cite{c4, c5}. Nevertheless, for formation planning problems in three-dimensional (3D) space, there is no existing method to obtain a globally rigid graph with minimum edges \cite{c6, c7}. If connected edges are cut off arbitrarily, the formation performance will be weakened tremendously. Thus, how to construct a globally rigid sparse graph for formation planning remains an open problem.

The objective of this paper is to establish a good sparse graph, which is globally rigid and preserves the structural feature of complete graphs.
We transform the sparse graph construction into submatric selection \cite{c8} of Laplacian matrix.
Due to the NP-hard selection problem, we introduce a genetic algorithm and the Max-Trace matrix-revealing metric to facilitate the solution.
The computation complex of sparse-graph-enabled formation planning is analyzed. 
Simulation, benchmark, and ablation study are performed. We provide preliminary advice on how to choose a reasonable rate of connection according to the test results.
The contributions of this paper are as follows:
\begin{enumerate}
    \item We design a graph sparsification mechanism and prove that the sparsified graph is globally rigid, which is a necessary condition to form a specific formation.
    \item We propose a good sparse graph construction method by submatrix selection to capture the predominant feature of the corresponding complete graph. 
    \item We integrate sparse graphs into formation trajectory planning to replace the computationally demanding complete graph. Extensive tests are conducted to validate that our method can well balance the computation efficiency and formation performance.
\end{enumerate}

\section{Related Works}

\subsection{Formation Trajectory Planning}
% Formation flight is the fundamental capability for swarms to execute cooperative task, and extensive works are dedicated to achieving this goal\cite{c9}. Morgan et al. \cite{c17} present a distributed planning method to guide hundreds to thousands of robots to form a desired shape, but the formation constraints are not considered during the flight process. Alonso-Mora et al. \cite{c16} use efficient sequential convex optimization to achieve formation planning of 16 quadrotors, whereas formation keeping constraints are ignored in local planning. Zhou et al. \cite{c18} design a swarm system with autonomous micro drones, the formation flight experiments are implemented with 10 drones. Especially, the relative position relation among drones is used to formulate the cost term in the trajectory planning problem to satisfy the formation constraints. Although these efforts of obstacle avoidance within formations, it is still challenging to achieve a larger-scale formation in dense environments.

Formation flight is the fundamental capability for swarms to execute cooperative task, and extensive works are dedicated to achieve this goal\cite{c9}. Kushleyev et al. \cite{c15} develop a centralized formation planning method, and the formation flight experiment with 20 micro-UAVs and motion capture system is presented.
Morgan et al. \cite{c17} present a decentralized planning method to guide hundreds to thousands of robots to form a desired shape, whereas formation keeping constraints are not considered during the flight process. Alonso-Mora et al. \cite{c16} use efficient sequential convex optimization to achieve formation planning of 16 quadrotors, but the formation performance reduces obviously in cluttered environments. Zhou et al. \cite{c18} design a fully autonomous swarm system with micro drones, the formation flight experiments are implemented with 10 drones in the wild. The relative position among drones is adopted to keep simple formation configuration in the trajectory planning problem. Although efforts have been made to swarm formation flight, the abovementioned studies suffer from low flexibility and adaptability for complex formation requirements.

\subsection{Graph-Based Formation Navigation}
A number of researches finely model collaborative relationships using graphs to improve formation performance. Most of them\cite{c19,b2,c21,c23} focus on graph-based formation control. Anderson et al. \cite{c3} expound on the necessity of globally rigid graph for formation maintenance. Xiao et al. \cite{c22} use graphs to model the interaction topology among agents and enable formation control for up to 30 agents in obstacle-free environments. Falconi et al. \cite{c24} achieve graph-based formation control considering obstacle avoidance constraints, but only adopt simple dynamics of robots. By contrast, Quan et al. \cite{c1} present a formation trajectory planning method by exploiting the Laplacian matrix of complete graphs in cluttered environments while ensuring the high elasticity, flexibility, and adaptability of a formation. Furthermore, it can facilitate formation transformation and elastic deformation of swarms. Nevertheless, the high computational complexity restricts this method to tackle the formation planning problem in larger scale.

To alleviate the computational complexity, several works are conducted from the perspective of reducing the connected edges of complete graphs. Cheah et al. \cite{c25} establish a partial connection graph, in which the agents only connect with nearby ones, but this mechanism fails to capture the features of complete graphs and struggles to maintain formation in cluttered environments. Zhao et al. \cite{c27} achieve a balance between computational efficiency and performance in the field of visual simultaneous localization and mapping, which is relevant to the essential objective of our paper (i.e., efficiency-performance trade-off).
They employ submatrix selection to reduce the number of feature points while preserving the key features. Submatrix selection, as a commonly used method for dimensionality reduction, can better capture the features of original matrix, and has been adopted in various fields such as sensor selection and complex network analysis\cite{c26,c28,b1}. This paper attempts to introduce submatrix selection and construct a good sparse graph, intending to achieve high-performance and efficient formation trajectory planning for large-scale aerial swarms.

% By combining an appropriate submatrix metric and the combinatorial optimization algorithm, the required submatrix can be selected quickly \cite{b1, c28}. Therefore, a good sparse graph may be obtained via submatrix selection for large-scale swarm formations.

\begin{figure}[!t] %thpb
    \centering
    \includegraphics[width=3.2in]{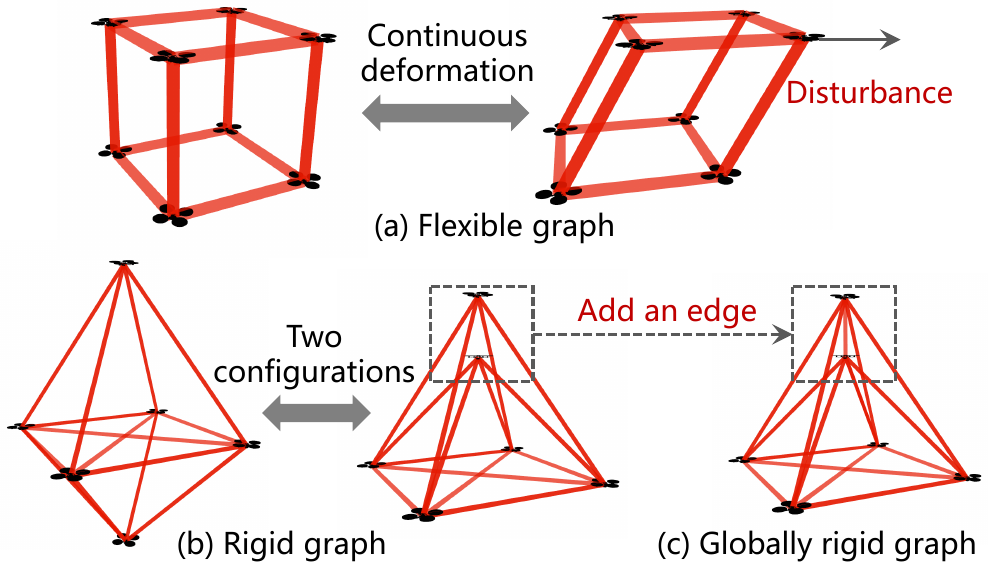}
    \caption{Illustration of graph rigidity. (a) A flexible graph deforms under disturbances. (b) A rigid graph corresponds to multiple geometric shapes. (c) A globally rigid graph ensures the stability and uniqueness.}
    \label{Graph_rigidity}
\end{figure}

\section{Preliminary and Overview}
% \subsection{Complete Graph Sparsification}
% \subsection{Good Sparse Graph Construction}
% \subsection{Sparse-graph-based Formation Planning}
% \subsection{Computation Complex Analysis}
\subsection{Graph-Based Formation and Laplacian Matrix}
The formation of $N$ drones is described by a directed graph $\mathcal G = (\boldsymbol v,\boldsymbol e)$, where $\boldsymbol v := {1, 2, ...,N}$ and $\boldsymbol e \subset \boldsymbol v \times \boldsymbol v$ represent the set of vertices and edges, respectively. 
In directed graph $\mathcal G$, the vertex $v_i \in \boldsymbol v$ denotes the $i$th drone with position vector $p_i = [x_i, y_i, z_i]^{\rm T}$. 
An edge $e_{ij} \in  \boldsymbol e$ connecting vertex $v_i$ and $v_j$ represents that only drone $i$ can obtain the geometric distance and trajectory information from drone $j$.
The weight of edge $e_{ij}$ is given by the Euclidean distance $w_{ij} =|| p_i-p_j ||_2$, where $||\cdot||_2$ denotes 2-norm.
Considering that graph theory is profound, we roughly introduce the basic concept of graph rigidity used in the following sections.
According to the definition in \cite{c29}, a globally rigid graph not only guarantees that the corresponding geometric shape does not undergo continuous deformation under disturbances, but also uniquely determines a shape. Therefore, the necessary condition for graph-based formation planning is that the graph $\mathcal G$ is globally rigid.
Fig. \ref{Graph_rigidity} depicts a graphical interpretation of graph rigidity.
\begin{figure}[!t]
    \centering
    \includegraphics[width=3.4in]{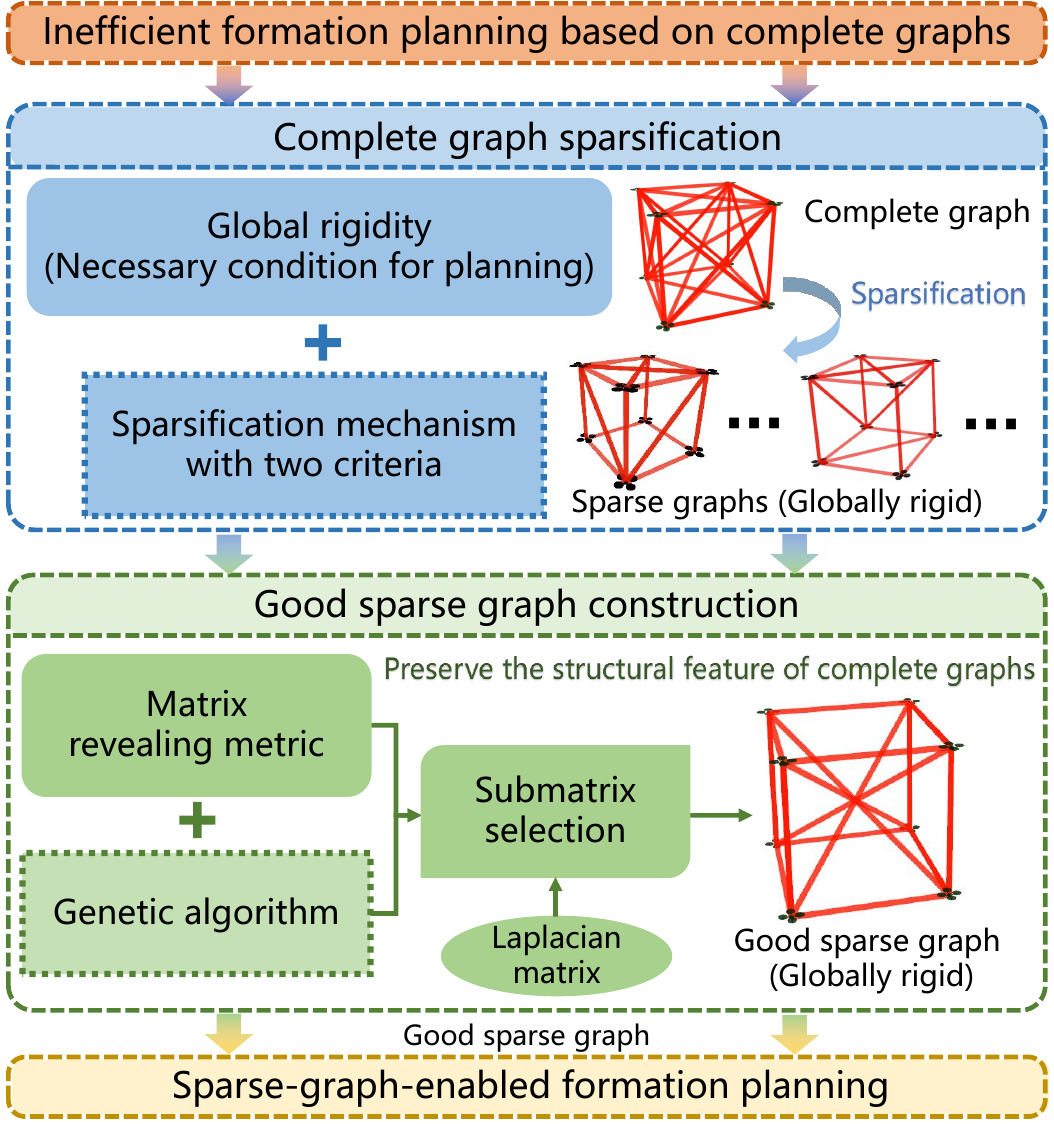}
    \caption{A diagram of our spare graph construct architecture.
    }
    \vspace{-0.2cm}
    \label{Framework}
\end{figure}
% In our previous work \cite{c1}, the formation graph is undirected and complete, which is globally rigid \cite{b3}.
% In contrast, this paper constructs sparse graphs, i.e., non-complete directed graphs, for formation planning to eliminate the high computational complexity.

Since Laplacian matrix $\mathbf L$ contains the structure information of graphs and reflects the connection relationship of vertices \cite{c30}, we use $\mathbf L$ to formulate the formation constraint.
\begin{equation}
    \mathbf L = \mathbf D - \mathbf A,
\end{equation}
where $\mathbf A\in \mathbb R^{N \times N}$ and $\mathbf D \in \mathbb R^{N \times N}$ denote the adjacency matrix and degree matrix of graphs, respectively.
The elements in $\mathbf A$ and $\mathbf D$ are obtained by 
\begin{equation}
\begin{aligned}
& A_{i,j} = \left\{
      \begin{aligned}
      &w_{ij}, &{\rm if} \, e_{ij} \in \boldsymbol e,\\
      &0,      &{\rm otherwise},\\
      \end{aligned}
      \right.\\
& D_{i,j} = \left\{
      \begin{aligned}
      &\sum\nolimits_{j=1}^N A_{i,j}, &{\rm if} \, i=j,\\
      &0,      &{\rm otherwise}.\\
      \end{aligned}
      \right.
\end{aligned}
\end{equation}

\subsection{Overall Architecture}

\begin{figure}[!t] %thpb
    \centering
    \includegraphics[width=3.4in]{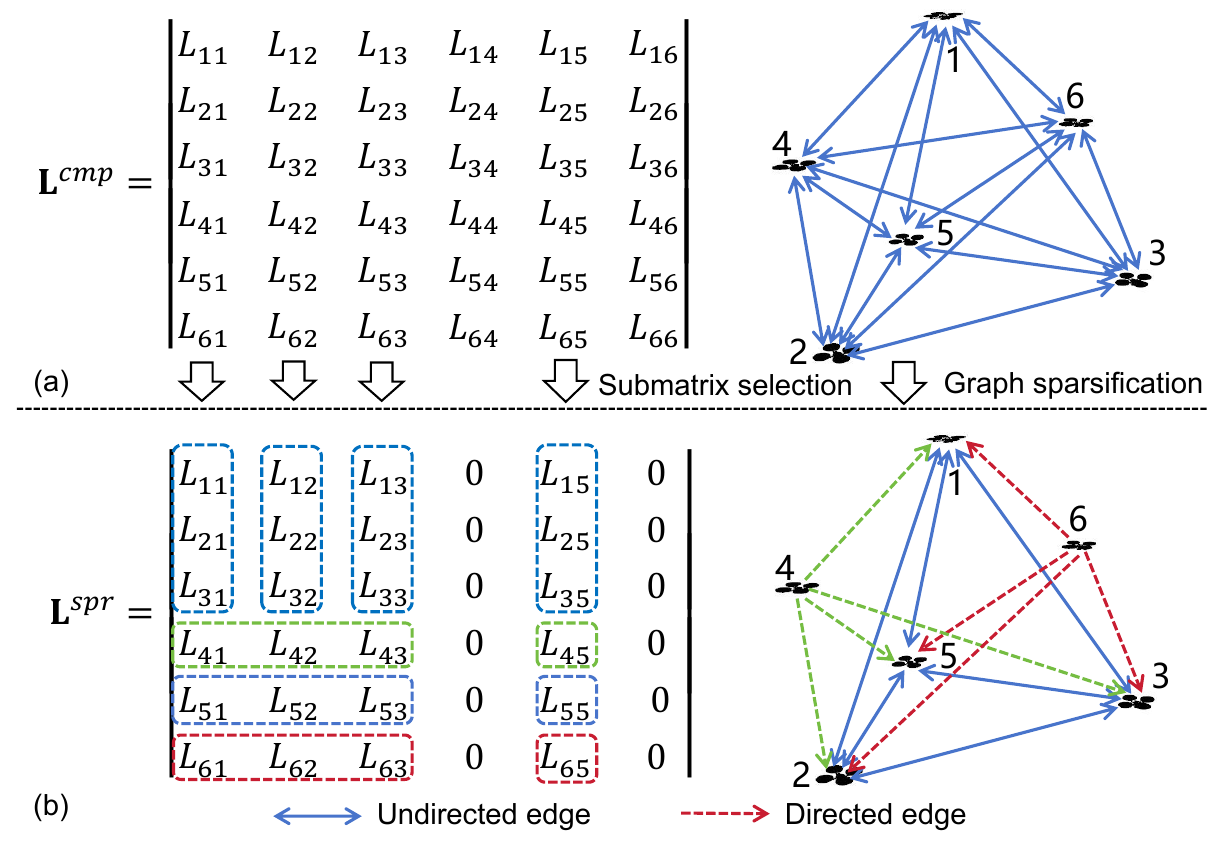}
    \caption{Illustration of submatric selection.
    (a) Laplacian matrix of complete graphs with six vertices.
    (b) Laplacian matrix of sparse graphs by submatrix selection, and the element in dashed box corresponds to the connected edge in the sparse graph with the same color.
    Four columns of the matrix are selected to form the submatrix, resulting in that vertices 1, 2, 3, \& 5 are selected as the base set $\boldsymbol v^{bas}$.
    Vertices 1, 2, 3, \& 5 connect with each other using undirected edges to establish a complete graph, and vertices 4 \& 6 connect with $\boldsymbol v^{bas}$ via directed edges.}
    \label{Submatrix}
\end{figure}

The overview of the sparse graph construction method is illustrated in Fig. \ref{Framework}.
In the second part (blue), the sparsification mechanism (Proposition 1 in Section IV-B) is designed as the foundation to ensure that the sparified graph is globally rigid
and corresponds to a geometric shape uniquely.
In the third part (green), a good sparse graph construction method is proposed to select key vertices from complete graphs to preserver the structural feature.
Since the graph is described by Laplacian matrix, choosing partial edges from complete graphs is equivalent to submatrix selection, i.e., extracting several columns from Laplacian matrix.
Fig. \ref{Submatrix} depicts a graphical interpretation of submatrix selection and graph sparsification.
The submatrix selection is formulated as a combinatorial optimization problem and is solved by matrix revealing metric and genetic algorithm.

% \begin{equation}
%     w_{ij} =|| p_i-p_j ||_2.
% \end{equation}

% Firstly, 
% Secondly, a good sparse graph construction method is proposed to select key vertices from complete graphs to preserver the structural feature.
% Lastly, the sparse graph is used in formation planning.

%Then $\mathbf A \in \mathbb R^{N \times N}$ and $\mathbf D \in \mathbb R^{N \times N}$ denote the adjacency matrix and degree matrix, and Laplacian matrix is given by
%\begin{equation}
%    \mathbf L = \mathbf D - \mathbf A.
%\end{equation}

% Choosing several columns of Laplacian matrix to form a submatrix is equivalent to sparsifying the connecting edges of a graph.  Fig. \ref{Submatrix} provides an example of submatrix selection. It is easy to know that the resulting construction mechanism for submatrix selection is identical. However, we cannot arbitrarily choose the columns of the matrix, even if the conditions of the \emph{Proposition} 1 are satisfied. In order to obtain a good submatrix, the selection process is formulated as a combinatorial optimization problem. By introducing matrix revealing metric to score submatrices and employing genetic algorithms to solve the submatrix selection problem, and metric is computed to capture the original matrix features. Through the remaining connections reflected by the selected submatrix, a good sparse graph is derived.

Lastly, the good sparse graph is used for formation planning.
Based on our previous work \cite{c1}, the formation planning is formulated as an unconstrained optimization problem
\begin{equation}
\rm P1: {\rm min} \, [\mathcal J_f, \mathcal J_o] \cdot \mu,
\end{equation}
% \mathop {\rm min}_{\textbf q, \textbf T} \, [\mathcal J_e, \mathcal J_t, \mathcal J_o, \mathcal J_f, \mathcal J_c, \mathcal J_d, \mathcal J_u] \cdot \mu,  
where $\mu$ denotes the weight vector. $\mathcal J_o$ contains the cost function of control effort, flight time, collision avoidance, and dynamical feasibility. 
$\mathcal J_f=f(\mathcal F_f)$ represents the cost function of swarm formation similarity, and $f(\cdot)$ is the differentiable metric to quantify the similarity distance $\mathcal F_f=||\mathbf L^{sqr}-\mathbf L^{sqr}_{des}||_F^2$ between current and desired formations. $\mathbf L^{sqr}$ is Laplacian matrix representing sparse graph $\mathcal G^{spr}$, and $\mathbf L^{sqr}_{des}$ denotes the matrix describing the desired formation configuration. $||\cdot||_F$ is Frobenius norm.

To analysis the computation complexity of P1, we derive the gradient of $\mathcal F_f$, since the cost of swarm formation similarity dominates the computationally expensive element.
According to the chain rule, the gradient of $\mathcal F_f$ with respect to the position of drones $p_i(t)$ can be obtained by
\begin{equation}
\frac{\partial{\mathcal F_f}}{\partial{p_i(t)}} =\frac{\partial{\mathcal F_f}}{ \partial{\mathbf w^{\rm T}_i}} \frac{\partial{\mathbf w^{\rm T}_i}}{\partial{p_i(t)}},
\end{equation}
where
\begin{equation}
    \frac{\partial{\mathcal F_f}}{ \partial{\mathbf w^{\rm T}_i}}=[\frac{\partial{\mathcal F_f}}{ \partial{\mathbf w_{i1}}},...,\frac{\partial{\mathcal F_f}}{ \partial{\mathbf w_{ij}}},...,\frac{\partial{\mathcal F_f}}{ \partial{\mathbf w_{in}}}],\ j \in \mathcal N_i,
\end{equation}
\begin{equation}
    \frac{\partial{\mathbf w^{\rm T}_i}}{\partial{p_i(t)}}=[\frac{\partial{\mathbf w_{i1}}}{\partial{p_i(t)}},...,\frac{\partial{\mathbf w_{ij}}}{\partial{p_i(t)}},...,\frac{\partial{\mathbf w_{in}}}{\partial{p_i(t)}}],\ j \in \mathcal N_i.
\end{equation}
$\mathcal N_i$ is the adjacent vertex set of drone $i$, and ${\mathbf w^{\rm T}_i}$ is a weight vector composed of edge weights in $\mathcal N_i$.
According to (5) \& (6), the computation complexity of $\partial{\mathcal F_f} / \partial{\mathbf w^{\rm T}_i}$ and $\partial{\mathbf w^{\rm T}_i}/\partial{p_i(t)}$ is $\mathcal O(N)$ when using complete graphs, i.e, the number of drones in $\mathcal N_i$ is $N-1$. Thus, $\partial{\mathcal F_f}/ \partial{p_i(t)}$ consumes $\mathcal O(N^2)$.
For sparse-graph-enabled formation planning, the computation cost is reduced to $\mathcal O((\varrho_cN)^2)$ with the connection rate $\varrho_c \in (0,100\%)$.
If $\varrho_c$ is set as 30\%, the complexity is almost one order of magnitude lower than that of complete graphs.
The simulation results in Fig. \ref{Time_edist_compar} also demonstrate the efficiency advantage of our method.
For the cost $\mathcal J_o$, metric $f(\cdot)$, and the solution of P1, readers can refer to \cite{c1} for more details.

% Based on our previous work \cite{c1}, the formation planning is formulated as an unconstrained optimization problem
% \begin{equation}
%    \mathop {\rm min}_{\textbf q, \textbf T} \, [\mathcal J_e, \mathcal J_t, \mathcal J_o, \mathcal J_f, \mathcal J_c, \mathcal J_d, \mathcal J_u] \cdot \mu,  
% \end{equation}
% where $\mu$ denotes the weight vector. $\mathcal J_e, \mathcal J_t, \mathcal J_o, \mathcal J_f, \mathcal J_c, \mathcal J_d$, and $\mathcal J_u$ represent the cost function of control effort, flight time, obstacle avoidance, swarm formation similarity, inter-drone collision avoidance, dynamical feasibility, and uniform distribution of constraint points, respectively.
% MINCO is adopted to represent the flight trajectory, and the optimization variables $\textbf q, \textbf T$ are the parameters of the MINCO trajectory, i.e., a series of intermediate points and the corresponding time vector. Please refer to \cite{c32} for more details of the MINCO trajectory representation, cost functions, and gradients of cost. L-BFGS is used to solve this optimization problem.

% As for the swarm formation similarity cost $\mathcal J_f$, the sparse graph $\mathcal G^{spr}$ is used to construct the Laplacian matrix $\mathbf L^{spr}$.
% The differentiable metric $\mathcal F_f$ quantifying the similarity distance between current formation and desired formation is given as
% \begin{equation}
%     \mathcal F_f=||\mathbf L^{sqr}-\mathbf L^{sqr}_{des}||_F^2,
% \end{equation}
% where $||\cdot||_F$ is Frobenius norm. 

\section{Sparsification of Complete Graphs}

For a complete graph $\mathcal G^{cmp}=(\boldsymbol v^{cmp}, \boldsymbol e^{cmp})$, we propose a graph sparsification mechanism to construct a corresponding sparse graph $\mathcal G^{spr}=(\boldsymbol v^{spr}, \boldsymbol e^{spr})$ that is also globally rigid, where $\boldsymbol v^{spr}=\boldsymbol v^{cmp}:={1, 2,..., N}$ and $\boldsymbol e^{spr} \subset \boldsymbol e^{cmp}$.

\emph{Proposition} 1 (Sparsification Mechanism): Given a set of vertices $\boldsymbol v^{spr}:=\{1, 2,...,N\}$, a sparse graph $\mathcal G^{spr}$ that satisfies the globally rigid can be constructed, if the following criteria are satisfied.
\begin{enumerate}[i)]
\item Select several vertices from $\boldsymbol v^{spr}$ and form the base set $\boldsymbol v^{bas}$. The number of vertices in $\boldsymbol v^{bas}$ is greater than or equal to 4, and these vertices should be non-coplanar in 3D space. Then, use $\boldsymbol v^{bas}$ to construct a complete graph $\mathcal G^{bas}=(\boldsymbol v^{bas}, \boldsymbol e^{bas})$.
\item Set the remaining vertices as $\boldsymbol v^{rmn}=\boldsymbol v^{spr}-\boldsymbol v^{bas}$. Connect each vertex in $\boldsymbol v^{rmn}$ to all vertices in $\boldsymbol v^{bas}$, but the vertices in $\boldsymbol v^{rmn}$ do not connect with each other.
\end{enumerate}

% In order to prove the sparsification Mechanism, we introduce there relevant Lemmas first, as shown in the Appendix.

\emph{Proof}: From the first criterion in the mechanism, the number of vertices in the base set $^{l}|\boldsymbol v^{bas}| \geq 4$, where $^{l}|\cdot|$ is used to extract the length of a vector.
The corresponding graph $\mathcal G^{bas}$ is a complete graph and is globally rigid.
Then, select a vertex $\boldsymbol v^{rmn}_i$ from the remaining vertex set $\boldsymbol v^{rmn}$ and connect it to all the vertices in $\boldsymbol v^{bas}$. Let $\mathcal G^{rmn}_i$ denotes the newly formed graph involving $\boldsymbol v^{rmn}_i$ and $\boldsymbol v^{bas}$.
Since $\mathcal G^{rmn}_i$ satisfies the conditions in Lemmas 1 \& 2,  $\mathcal G^{rmn}_i$ is globally rigid in 3D space.

\emph{Lemma} 1 (\emph{Extension lemma}) \cite{c7}. Let $\mathcal G_1$ be a graph obtained from a graph $\mathcal G_2$ by adding a new vertex $v$ with $k$ edges incident to $v$.
If $\mathcal G_2$ is globally rigid in $\mathbb R^d$ and $k \geq d+1$, then  $\mathcal G_1$ is globally rigid in $\mathbb R^d$.

\emph{Lemma} 2 \cite{c33}. A graph is globally rigid if it is formed by starting from a clique of four non-coplanar nodes and repeatedly adding a node connected to at least four non-coplanar existing nodes.

Assume that there exist $m$ vertices in $\boldsymbol v^{rmn}$, according to the second criterion in the mechanism, the corresponding $m$ graphs $\mathcal G^{rmn}_1,..., \mathcal G^{rmn}_i,..., \mathcal G^{rmn}_m$ that are globally rigid can be constructed.
Then, define the sparse graph $\mathcal G^{spr}=\mathcal G^{rmn}_1 \cup,..., \mathcal G^{rmn}_i \cup, ..., \cup \mathcal G^{rmn}_m$, i.e., the $\mathcal G^{spr}$ is formed after all vertices in $\boldsymbol v^{rmn}$ are connected to all vertices in $\boldsymbol v^{bas}$. Note that the vertices in $\boldsymbol v^{rmn}$ are not connected with each other. It can be known that $\mathcal G^{rmn}_1 \cap,..., \mathcal G^{rmn}_i \cap, ..., \cap \mathcal G^{rmn}_m = \mathcal G^{bas}$.
Due to that $^{l}|\boldsymbol v^{bas}| \geq 4$ and all $\mathcal G^{rmn}_i$ are globally rigid, the constructed sparse graph satisfies the conditions in Lemma 3. Thus,  $\mathcal G^{spr}$ is a globally rigid graph.

\emph{Lemma} 3 (\emph{Gluing lemma}) \cite{c7}. Let $\mathcal G_1$ and $\mathcal G_2$ be graphs with $|V (\mathcal G_1) \cap V (\mathcal G_2)| = k$, and let $\mathcal G =  \mathcal G_1 \cup \mathcal G_2$. If $\mathcal G_1$ and $\mathcal G_2$ are globally rigid in Rd and $k \geq d+1$, then $\mathcal G$ is globally rigid in $\mathbb R^d$. $V(\mathcal G)$ represents the set of vertices in graph $\mathcal G$, and $|V(\mathcal G)|$ denotes the number of vertices. \hfill $\blacksquare$

Through the above construction mechanism, a sparse graph with the same vertices in complete graphs can be obtained. Considering the vertices as drones, it can be observed that the number of position information that needs to be taken into account for each drone is only $^{l}|\boldsymbol v^{bas}|$ and less than $^{l}|\boldsymbol v^{bas}|+^{l}|\boldsymbol v^{rmn}|-1$, thereby reducing the number of formation constraints.

\section{Good Sparse Graph Construction Method}

\subsection{Good Sparse Graph Construction Problem} 

According to the sparsification mechanism in Section IV, $\mathcal G^{bas}$ is essential for sparse graphs, since 
$\mathcal G^{bas}$ can be regraded as a leader including a group of vertices intuitively. 
In order to guarantee the formation performance, $\mathcal G^{bas}$ should be selected finely to construct a good sparse graph, which can capture the main feature of original complete graphs sufficiently. 

Define the good sparse graph construction problem to be: given a desired formation, find a sparse graph with only a part of edges, such that the formation error of swarm trajectories is minimized when using the sparse graph.

We use the Laplacian matrix of sparse graphs $\mathbf L^{spr} \in \mathbb R^{N \times N}$ to describe the formation constraints in the planning problem.
The sparse graph construction can be transformed into the submatrix selection, which is formulated as the following combinatorial optimization problem:
% \begin{equation}
% \begin{aligned}
% & \underset{H^{clm}}{\text{min}} \, ||\mathbf L^{cmp}-\mathbf L_{[H^{clm}]}||_2,\\
% &s.t.\ \
% H^{clm} \subseteq \{1,2,...,N\},\\
% &\ \ ^{l}|H^{clm}| = ^{l}|\boldsymbol v^{bas}|,\\
% & ^{l}|\boldsymbol v^{bas}| \geq 4, \\
% & \text{all vertexes in } \boldsymbol v^{bas} \text{ are non-coplanar},\\
% \end{aligned}
% \end{equation}

\begin{equation}
    \begin{aligned}
    % \underset{H^{clm}}{\text{min}} \, ||\mathbf L^{cmp}-\mathbf L_{[H^{clm}]}||_2,\\
    {\rm P}2: \min_{H^{clm}} & \,||\mathbf L^{cmp}-\mathbf L^{cmp}_{[H^{clm}]}||_2,\\
    \end{aligned}
\end{equation}
\begin{equation}
    \begin{aligned}
    % \underset{H^{clm}}{\text{min}} \, ||\mathbf L^{cmp}-\mathbf L_{[H^{clm}]}||_2,\\
    \text{s.t.}\ \
    & H^{clm} \subseteq \{1,2,...,N\},\\
    & ^{l}|H^{clm}| = ^{l}|\boldsymbol v^{bas}| \geq 4,\\
    % & \text{all vertices in } \boldsymbol v^{bas} \text{ are non-coplanar},\\
    &\boldsymbol v^{bas}\ \text{is non-coplanar},\\
    \end{aligned}
\end{equation}
where $H^{clm}$ is the optimization variable and contains the index subsets of selected column blocks from Laplacian matrix $\mathbf L^{cmp}$.
$\mathbf L^{cmp}_{[H^{clm}]}=\mathbf L^{spr} \in \mathbb R^{N \times N}$ is the column-wise selected submatrix, i.e., only the column blocks with the index in $H^{clm}$ have elements that are identical with that in $\mathbf L^{cmp}$, and the rest of the column blocks are $\mathbf 0 \in \mathbb R^{N \times 1}$.

The objective of submatrix selection $||\mathbf L^{cmp}-\mathbf L^{cmp}_{[H^{clm}]}||_2$ attempts to preserve the structural information of the original complete graph. We find that the sparse graph constructed by this method can guarantee promising performance in formation planning, and verify this method through simulation in Section VI. The better $\mathbf L^{spr}$ is selected, the better $\boldsymbol v^{bas}$ can be determined.
Meanwhile, the constraints in (8) ensure that $\boldsymbol v^{bas}$ satisfies the criteria in sparsification mechanism, thus the formed sparse graph is globally rigid.
If the selected vertices in $\boldsymbol v^{bas}$ is coplanar, a new vertex will be added in $\boldsymbol v^{bas}$ until the third constraint in (8) is satisfied.

\subsection{Matrix-Revealing-Metric-Based Submatrix Selection}
Although the combinatorial optimization problem of submatrix selection in ${\rm P}2$ can be solved by enumeration, the exponentially-growing problem dimension encounters combinatorial explosion for large-scale formation. To enhance the efficiency of selection and confine the loss in optimality, the genetic algorithm \cite{c31} is introduced, and we finely tune the size of population, number of generation, and probabilities of exchange crossover and mutation to balance the performance and computation efficiency.
% \makeatletter
% \newcommand{\removelatexerror}{\let\@latex@error\@gobble}
% \makeatother
% \documentclass{article}
% \usepackage{algorithm}
% \usepackage{algpseudocode}
% \usepackage{amsmath}
% \begin{figure}[!t]
%   \begin{algorithm}[H]
%     \caption{Good Sparse Graph Construction}
%     \textbf{Notation:} The sequence number of the columns selected after Submatrix\_Selection $H^{clm*}$, The sequence number of the columns of $\mathbf L^{cmp}$ (except $H^{clm*}$) $X^{rmn}$, Selected four non\_coplanar points $v^{4nc}$\\
%     $\mathbf L^{cmp} \leftarrow \{\boldsymbol v^{cmp},\boldsymbol e^{cmp}\}$\\
%     $\mathbf L^{spr}, H^{clm*} \leftarrow \mathbf{Submatrix}(\mathbf L^{des})$\\
%     $v^{bas} \leftarrow \mathbf{corresponding\_vertexes}(H^{clm*})$\\
%     \If{$\mathbf{Is\_coplanar}(v^{bas})$}{
%         $v^{4nc} \leftarrow \mathbf{Active\_selection}(v^{cmp})$\\
%         $\mathbf L^{spr}, H^{clm*} \leftarrow \mathbf{Submatrix}(\mathbf L^{des}, v^{4nc})$\\
%     }
%     \Else{
%         $\mathcal{G}^{bas} \leftarrow \mathbf{Complete\_Graph}(v^{4nc},H^{clm*})$\\
%         $\mathcal{G}^{spr} \leftarrow \mathbf{Remainder\_Connect}(X^{rmn},\mathcal{G}^{bas})$\\
%     }
%     Formation\_Planning($\mathcal{G}^{spr}$) \hfill $\triangleright$ detailed in Sec.\uppercase\expandafter{\romannumeral5}
%   \end{algorithm}
% \end{figure}

\begin{figure}[!t] %thpb
    \centering
    \includegraphics[width=3.2in]{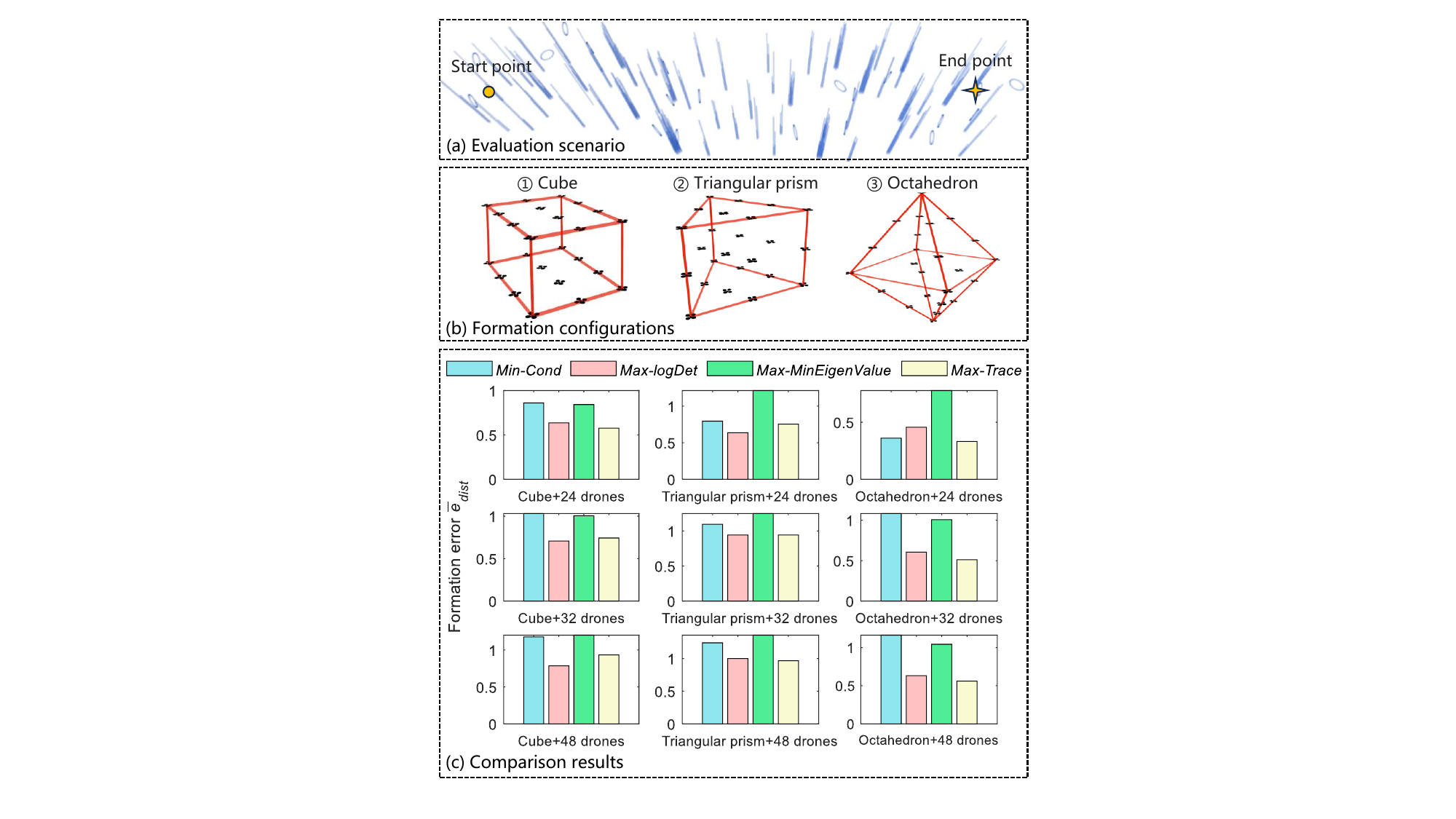}
    \vspace{-0.1cm}
    \caption{Comparative simulation on four candidate matrix-revealing metrics. 
    (a) We simulate drones flying in formation from left side in a cluttered map to right side with a velocity limit of 2m/s. The map is generated randomly.
    The global formation trajectory is planned from the start point to end point.
    (b) Three formation configurations are tested. 
    (c) Comparison results of formation error $\bar e_{dist}$ along the global trajectory are provided.}
    \label{Matrix_score}
\end{figure}
The computation complexity of the objective function in ${\rm P}2$ is $\mathcal O(N^3)$, which is impractical for the solution of combinatorial optimization.
Inspired by \cite{c27}, the matrix-revealing metric is adopted to replace the time-cost objective function, and the optimization problem in ${\rm P}2$ can be converted into
% \begin{equation}
% \begin{split}
% &\mathop {\rm max}_{H^{clm} \subseteq \{1,2,...,N\}} \, \mathcal R(\mathbf L_{[H^{clm}]}),\\
% &s.t.\quad  \left\{\begin{array}{lc}
% ^{l}|H^{clm}| \geq 4,\\
% {All \ vertexes} \in \boldsymbol v^{bas} {\ are} {\ non} {-} {coplanar},\\
% \end{array}\right.
% \end{split}
% \end{equation}

\begin{equation}
    \begin{aligned}
    % \underset{H^{clm}}{\text{min}} \, \mathbf L^{cmp}-\mathbf L_{[H^{clm}]}||_2,\\
   {\rm P}3: \min_{H^{clm}} & \, \mathcal R(\mathbf L^{cmp}_{[H^{clm}]}),\\
    \text{s.t.}\ \
    & (8),\\
    \end{aligned}
\end{equation}
where $\mathcal R(\cdot)$ is the matrix-revealing metric.
There exist several commonly used metrics, and we conduct simulation to assess four candidates listed in Table I to explore which one performs best to assist submetrix selection.
\begin{table}[thpb]
  \renewcommand{\arraystretch}{1.2}
  \caption{Candidate matrix-revealing metrics}
  \label{tab1}
  %\footnotesize
  \tabcolsep 3.5pt %space between two columns.
  \centering
  \begin{threeparttable}
    \begin{tabular}{ll}%{\textwidth}
      \toprule
      \makecell[l]{\emph{Min-Cond}\cite{c27}} &\makecell[l]{Minimize condition $\lambda_{\rm max}(\mathbf L)/ \lambda_{\rm min}(\mathbf L)$}  \\
      \makecell[l]{\emph{Max-logDet}\cite{c26}} &\makecell[l]{Maximize Log of determinant ${\rm log}({\rm det}(\mathbf L))$} \\
      \makecell[l]{\emph{Max-MinEigenValue}\cite{c28}} &\makecell[l]{Maximize minimal eigenvalue $\lambda_{\rm min}(\mathbf L)$} \\
      \makecell[l]{\emph{Max-Trace}\cite{b1}} &\makecell[l]{Maximize trace $Tr(\mathbf L)=\sum_1^NL_{ii}$} \\
      \bottomrule
    \end{tabular}
    \begin{tablenotes}
      \footnotesize
      \item $\lambda_{\rm min}$ \& $\lambda_{\rm max}$ are minimal \& maximal eigenvalues of square matrix.
    \end{tablenotes}
  \end{threeparttable}
\end{table}

To evaluate these metrics soundly, we simulate three formation configurations (named as cube, triangular prism, and octahedron) in a cluttered environment of 60m $\times$ 20m size, as illustrated in Figs. \ref{Matrix_score}(a)\&\ref{Matrix_score}(b).
The numbers of drones 24, 36, and 48 are tested for each configuration. For genetic algorithm, the size of population and number of generation are set as 6000 and 100. The probabilities of exchange crossover and mutation are 0.4 and 0.4. The simulation platform is a personal computer with an Intel Core i7 8700K CPU running at 3.2 GHz and with 32-GB RAM at 3200 MHz.
We implement our proposed method in C++.

The sparse graphs obtained by different matrix-revealing metrics are employed in formation planning.
We slightly modify the measurement method in \cite{c1} to acquire the average formation error $\bar e_{dist}$ as the quantitative indicator.
\begin{equation}
    \begin{aligned}
    &\bar e_{dist}= \frac{1}{L^{fma}_{\rm max}L^{trj}} \int_{\mathcal L} \mathop{\rm min}_{\mathbf R, \mathbf t, s} \sum_{i=1}^N ||\mathbf p_i^{des} - (s \mathbf R \mathbf p_i^{atu}+\mathbf t)||_2 dl,\\
    \end{aligned}
\end{equation}
where $L^{fma}_{\rm max}$ and $L^{trj}$ are the maximum diagonal length in the formation and the length of swarm trajectory $\mathcal L$, respectively.
The combination of rotation $\mathbf R \in SO(3)$, translation $\mathbf t \in \mathbb R^3$, and scale expansion $s \in \mathbb R_+$ represent the similarity transformation that aligns actual flight formation $\mathcal A^{atu}$ with desired one $\mathcal A^{des}$ to normalize the formation configuration.
$\mathbf p_i^{des}$ and $\mathbf p_i^{atu}$ denote the position of the $i\rm{th}$ drone in $\mathcal A^{atu}$ and $\mathcal A^{des}$.
Since the influence of rotation, translation, scaling, and length of trajectory is eliminated by normalization, $\bar e_{dist}$ can fairly measure the distortion degree between $\mathcal A^{atu}$ and  $\mathcal A^{des}$ along the global formation trajectories. 

The comparison results among different matrix-revealing metrics are provided in Fig. \ref{Matrix_score}(c). 
The $\bar e_{dist}$ by \emph{Max-Trace} and \emph{Max-logDet} is comparative and outperforms the other two obviously.
The computation complexity of Max-Trace is only $\mathcal O(N)$ , whereas for \emph{Max-logDet} the cost is $\mathcal O(N^3)$ \cite{c27}.
Thus, \emph{Max-Trace} is chosen as the scoring metric to guide good submatrix selection.

% \section{Sparse-Graph-Enabled Formation Planning}

\section{Simulation and Benchmark}
\begin{figure}[!t]
    \centering
    \includegraphics[width=3.2in]{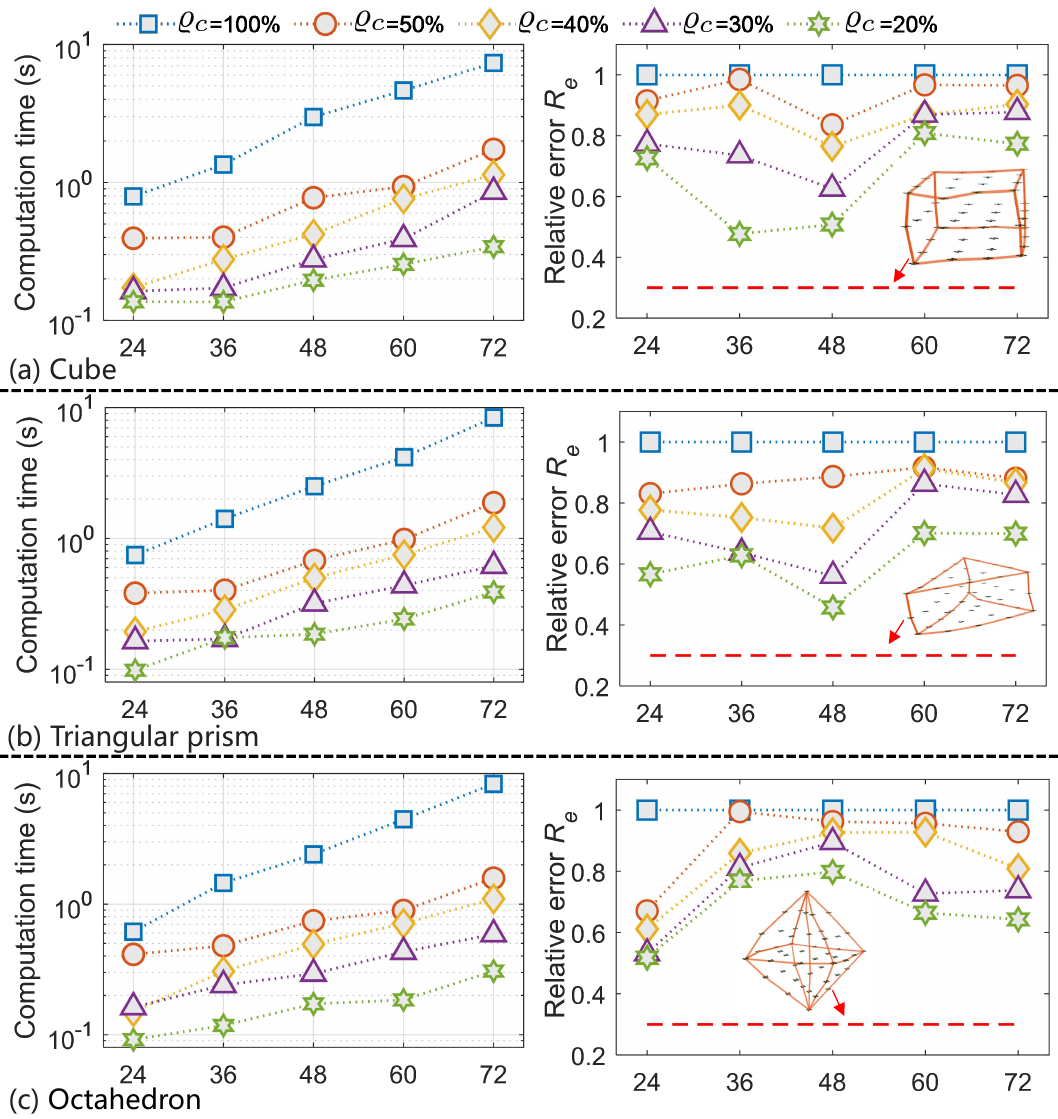}
    \vspace{-0.1cm}
    \caption{Comparison results of planning efficiency and relative formation error.
    (a)-(c) represent the results on different formation shapes (cube, triangular prism, and octahedron). 
    The connection rates of graphs $\varrho_c$ ranging from 20\% to 50\% are tested.
    The number of drones is set as 24, 36, 48, 60, and 72.
    The first and second columns of the figure represent the computing time and relative formation error, respectively.
    The data in the figure are the average value for 20 runs, and the map are randomly regenerated in each test.
    We also provide the conservative lower bound of relative formation error (red dashed line), under which the error becomes unacceptable.}
    \vspace{-0.2cm}
    \label{Time_edist_compar}
\end{figure}
\begin{figure}[!t]
    \centering
    \includegraphics[width=3.1in]{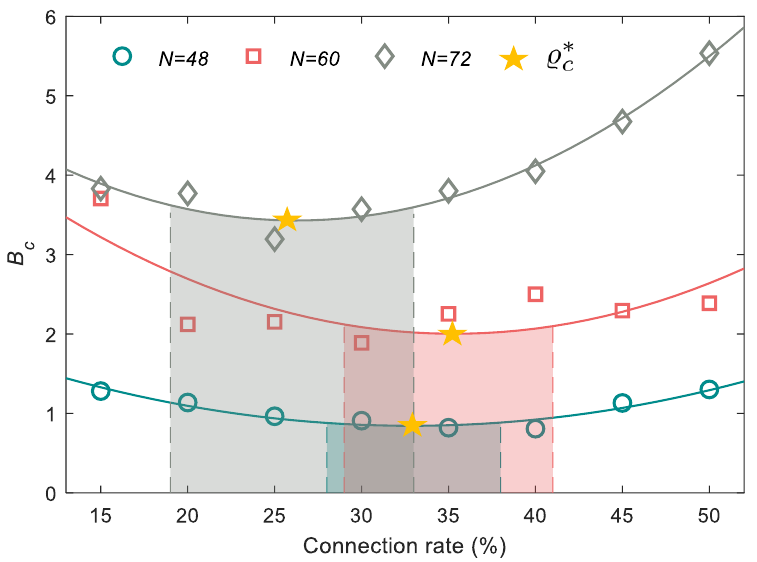}
    \caption{Analysis of planning efficiency and formation error. The shaded area indicates the favorable interval of $\varrho_c$ to balance $t^{cpu}$ and $\bar e_{dist}$.
    The minimal point on each curve is $\varrho^*_c$, then the range of the shaded area is [$\varrho^l_c$, $\varrho^r_c$], where $B_c(\varrho^l_c) = B_c(\varrho^r_c) = (1+5\%)B_c(\varrho^*_c)$.
    %The range of the shaded area is $\pm$5\% of the minimal point on each curve.
    }
    \label{Fitcuver}
\end{figure}

\subsection{Study of Planning Efficiency and Performance}

This subsection focuses on analyzing the computation time used for planning and formation performance with different connection rates of graphs and varying number of drones.
The hardware and simulation environments are set as the same in Section V-B. 
For the study of formation performance, we regard the average formation error of complete-graph-based planning $\bar e_{dist}(100\%)$ as the baseline,
and design the relative formation error in (11) to measure how close the error by sparse graphs is to the baseline.
\begin{equation}
    R_e= 1-\frac{\bar e_{dist}(\varrho_c)-\bar e_{dist}(100\%)}{\bar e_{dist}(100\%)},
\end{equation}
where $\bar e_{dist}(\varrho_c)$ denotes the average formation error under connection rate $\varrho_c$, and $R_e=1$ when $\varrho_c=100\%$.
The closer $R_e$ is to 1, the better formation performance is obtained. 

Fig. \ref{Time_edist_compar} depicts the computation time and relative formation error $R_e$ w.r.t. complete graphs.
From the results, we observe that the computation time reduces considerably as $\varrho_c$ decreases across various test scenarios.
For 72 drones, the computation efficient under $\varrho_c$ = 30\% is over $10\times$ faster than the counterpart with $\varrho_c$ = 100\%.
Although the method suffers from performance drops as $\varrho_c$ decreases,  $R_e$ is still larger than the conservative lower bound even when $\varrho_c$ = 20\%.
As a result, the real-time formation planning with 72 drones can be achieved when $\varrho_c$ is below 30\%, and comes an acceptable and adjustable sacrifice of formation error.
The formation planning under complete graphs ($\varrho_c$ = 100\%) costs seconds and becomes computationally prohibitive for real-time requirements.

To comprehensively evaluate the efficiency and performance and provide a preliminary suggestion on how to choose a reasonable $\varrho_c$,
we develop a trade-off formula
\begin{equation}
    B_c(\varrho_c)=(t^{cpu}(\varrho_c))^{\alpha} - \kappa \cdot (\bar e_{dist}(\varrho_c))^{\beta},
\end{equation}
where $\kappa=8\times10^{-6}$ is the weight coefficient, and $\alpha=4, \beta=1.75$ are exponential coefficients.
Several $t^{cpu}$ and $\bar e_{dist}$ are computed with $\varrho_c$ ranging from 15\% to 50 \% with the interval of 5\%.
Three numbers of drones 48, 60, and 72 are tested, and corresponding $t^{cpu}$ and $\bar e_{dist}$ are fed into (10).
Then, a series of $B_c$ are fitted as three quadratic curves in Fig. \ref{Fitcuver}.
$\varrho_c$ can be adjusted based on user-defined requirements, and we suggest to select $\varrho_c$ in [25\%, 35\%] to obtain a well compromise solution according to the evaluation results.

\subsection{Benchmark of Sparse-Graph-Based Formation Planning}
To demonstrate the superiority of our method, benchmark comparisons are performed.
In Table II, three sparse graph construction methods including random sparse graph (\emph{random}),  nearest-neighbor sparse graph (\emph{nearest}) \cite{c25}, and our method without the optimization on base set $\boldsymbol v^{bas}$ (\emph{ours w/o opt}) are compared with the proposed method in terms of formation error and formation recovery. The test of complete graphs (\emph{complete}) is also introduced as the reference.
For fair comparison, all the constructed graphs are integrated into the same formation planning problem in Section III-B.
\begin{table}[thpb]
  \renewcommand{\arraystretch}{1.3}
  \caption{Comparative methods in Benchmark}
  \label{tab2}
  %\footnotesize
  \setlength\tabcolsep{3.5pt} %space between two columns.
  \centering
  \begin{threeparttable}
    \begin{tabular}{ll}%{\textwidth}
      \toprule
      \makecell[l]{\emph{Random}} &\makecell[l]{Each drone randomly selects $^l|\boldsymbol v^{bas}|$ drones to connect}  \\
      \makecell[l]{\emph{Nearest}} &\makecell[l]{Each drone selects $^l|\boldsymbol v^{bas}|$ nearest drones to connect} \\
      \makecell[l]{\emph{Ours w/o opt}} &\makecell[l]{Base set $\boldsymbol v^{bas}$ is selected randomly without optimization} \\
      \makecell[l]{\emph{Complete}} &\makecell[l]{Each drone connects with all other drones} \\
      \bottomrule
    \end{tabular}
    \begin{tablenotes}
      \item \footnotesize $^l|\boldsymbol v^{bas}|$ is the number of connected edges of each drone.
    \end{tablenotes}
  \end{threeparttable}
\end{table}

\begin{figure}[!t]
    \centering
    \includegraphics[width=3.2in]{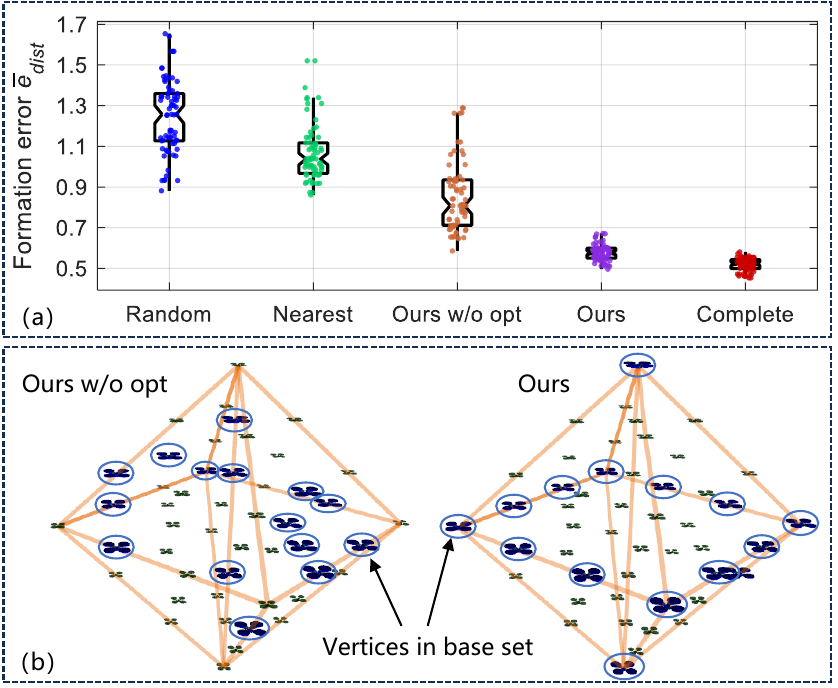}
    \caption{Comparative simulation on formation error.
    (a) Boxplot of formation error by different methods.
    (b) The visualization of base sets selected by \emph{ours w/o opt} and \emph{ours}, respectively.}
    \label{Benchmark_edist_obs}
\end{figure}
\begin{figure*}[!t]
    \centering
    \includegraphics[width=7.0in]{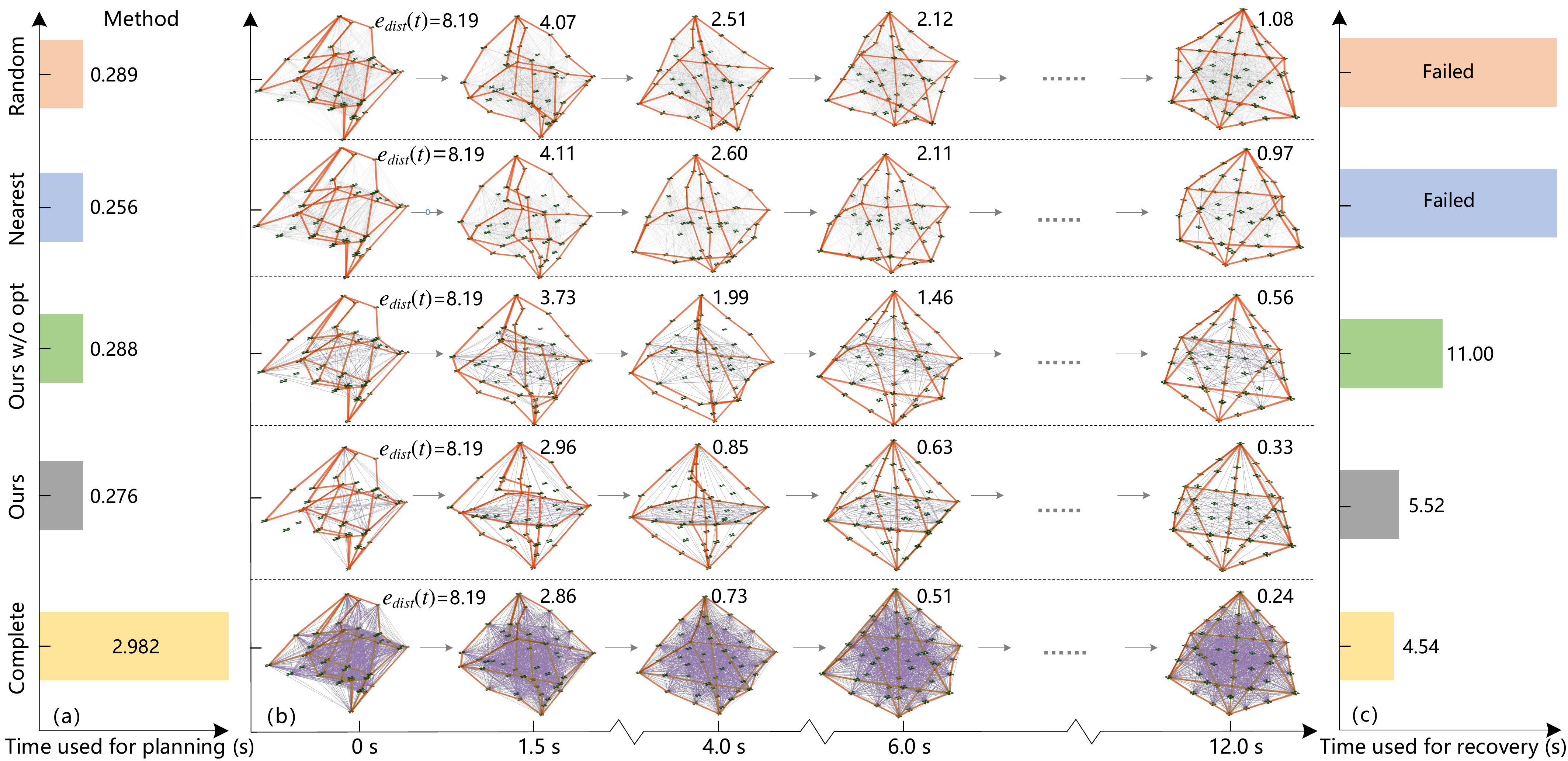}
    % \vspace{-0.1cm}
    \caption{Comparative simulation on formation recovery. 
    (a) The computation time for formation planning $t^{cpu}$.
    (b) The formation recovery process, and several formation states are provided at different time. The thick orange line represents the overall skeleton of the geometric shape.
    The thin grey line is the edge connecting different drones.
    $e_{dist}(t)=(\mathop{\rm min}_{\mathbf R, \mathbf t, s} \sum_{i=1}^N ||\mathbf p_i^{des} - (s \mathbf R \mathbf p_i^{atu}+\mathbf t)||_2)/L^{fma}_{\rm max}$ is the instantaneous formation error at time $t$.
    (c) The time used for formation recovery.
    If $e_{dist}(t) \leq 0.65$, it is consider that the formation is converged to the desired shape.}
    % \vspace{-0.4cm}
    \label{Recovery}
\end{figure*}

\begin{figure*}[!t]
    \centering
    \includegraphics[width=7.0in]{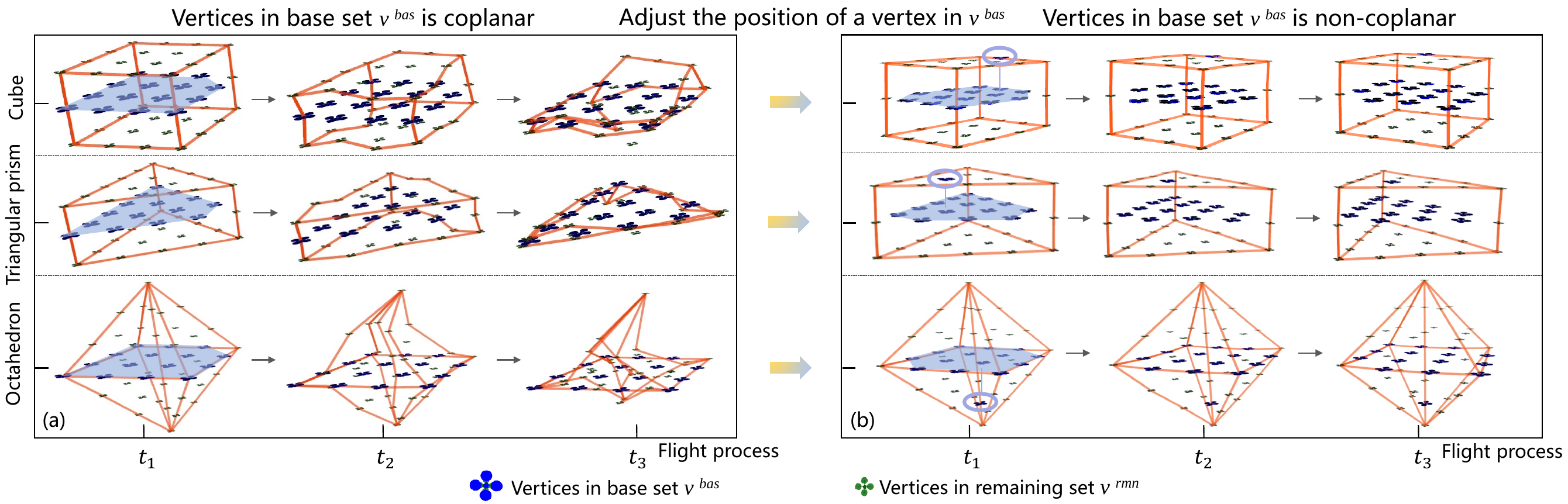}
    \caption{Formation planning results in the ablation study. 
    Three formation configurations are tested, and the number of drones and connection rate are set as 48 and 30\%, respectively.
    (a) The formation with coplanar vertices in base set $\boldsymbol v^{bas}$.
    The shaded area represents the vertices in $\boldsymbol v^{bas}$.
    (b) The formation with non-coplanar vertices in $\boldsymbol v^{bas}$.
    The vertices in the shaded area and circle represent $\boldsymbol v^{bas}$.}
    % The vertices in base set $\boldsymbol v^{bas}$ are larger than that in remaining set $\boldsymbol v^{rmn}$.}
    % \vspace{-0.5cm}
    \label{Ablation_study}
\end{figure*}

\subsubsection{Comparison on Formation Error}
The test scenario involving 48 drones is established to require keeping an octahedron formation.
The connection rate $\varrho_c$ is set as 30\%, thus the number of edges connected on each drones $^l|\boldsymbol v^{bas}|$ is set to 15.
Each method is run 80 times with randomly generated maps.
From the comparison results in Fig. \ref{Benchmark_edist_obs}(a), the average formation error $\bar e_{dist}$ and its dispersion of \emph{nearest} and \emph{random} are large.
\emph{Ours w/o opt} is better and benefits the global rigidity of the constructed sparse graph, but the randomly-selected base set make the formation performance unstable.
Our method outraces all the methods in Table II, and achieve lower $\bar e_{dist}$, with much tighter dispersion.
Besides, $\bar e_{dist}$ by ours is very close to that by \emph{complete}, which can be attributed in part to a better structural information captured by \emph{ours}. 
In Fig. \ref{Benchmark_edist_obs}(b), it can be found that the vertices in base set describe the skeleton of the graph more holistically than \emph{ours w/o opt}.

\subsubsection{Comparison on Formation Recovery}
A swarm with 48 drones is required to recovery from a scattered state to octahedron formation without obstacle avoidance, and the connection rate is set as 30\%.
From Fig. \ref{Recovery}(a), the sparse-graph-enabled formation planning highlights an order of magnitude’s speedup, w.r.t. \emph{complete}.
The formation recovery performance is qualitatively visible by plotting a subset of formation states in Fig. \ref{Recovery}(b).
We identify that ours can form the desired octahedron shape faster against the first three methods in Table II, and the time required for recovery in Fig. \ref{Recovery}(c) is very close to \emph{complete}.
In summary, the good sparse graph can facilitate the planing to obtain satisfactory formation results and improve the computation efficiency significantly.

\subsection{Ablation study of Graph Sparsification Mechanism}
%The Sparse Graph Construction method outlined in this paper specifies that for 3D formations, the points in $\boldsymbol v^{bas}$ must not be coplanar, which constitutes a crucial condition for ensuring the global rigidity of the graph. In order to demonstrate the necessity of these conditions and underscore the vital role of global rigidity in formation maintenance, we conducted tests on three different formations composed of 48 drones, each with 30\% sparse connections. However, all selected vertexes in $\boldsymbol v^{bas}$ were found to lie within a single plane (since three vertexes must always be coplanar, this specific scenario is not explicitly shown here), as depicted in Fig. \ref{Ablation_study}. During formation flight planning, only a segment corresponding to $\mathcal G^{bas}$ was able to maintain stable relative positions, while the remaining drones were entirely unable to do so. This occurs because those sparse graphs formed from coplanar vertexes lacks global rigidity, rendering it incapable of uniquely corresponding to a geometric shape, thus making it unsuitable for formation maintenance.%

To validate the key part of the graph sparsification mechanism in Section IV-B (i.e., vertices in $\boldsymbol v^{bas}$ should be non-coplanar in 3D), the ablation study is conducted.
As shown in Fig. \ref{Ablation_study}(a), we select coplanar vertices to form $\boldsymbol v^{bas}$ intentionally.
According to the graph rigidity theory in Section III-A, the constructed sparse graph with coplanar $\boldsymbol v^{bas}$ may be just a rigid graph instead of a globally rigid graph in 3D, which cannot determine a geometric shape uniquely.
Since different shapes correspond to different local minima and these local minima are close to each other, the formation planning oscillates among multiple local minima and cannot converge to a feasible solution, resulting in the collapse of formation. 
Additionally, we simply change the position of a vertex to establish a non-coplanar $\boldsymbol v^{bas}$ in Fig. \ref{Ablation_study}(b). It can be observed that the graph with non-coplanar $\boldsymbol v^{bas}$ can maintain stable formation for three configurations.

% It can be observed that the vertices in $\boldsymbol v^{bas}$ are able to maintain stable relative position, because $\boldsymbol v^{bas}$ construct a complete graph. However, the remaining vertices in $\boldsymbol v^{rmn}$ cannot keep the desired position and the 

% In the ablation study, coplanar vertices are selected to form $\boldsymbol v^{bas}$ intentionally. 
% \textcolor{red}{ As shown in Fig. 10(a), the vertices in $\boldsymbol v^{bas}$ is coplaner.  Therefore, $\boldsymbol v^{bas}$ must be non-coplanar to ensure the formation performance. By contrast, as shown in Fig. 10(b), we simplily change a vertex of $\boldsymbol v^{bas}$ in Fig. 10(a) so that it is non-coplanar, }

\section{CONCLUSIONS and FUTURE WORK}

In this paper, we propose a good sparse graph construction method for efficient formation planning. 
The graph sprasification mechanism and submetrix selection are integrated to ensure that the sprasified graph is globally rigid and captures the main structural feature of complete graphs.
Computation complex analysis and simulation with more than 50 drones demonstrate that the efficiency under sparse graphs preserving 30\% edges is higher than that under complete graphs by a factor of ten.
Benchmark comparisons and ablation studies indicate that our method can provide satisfactory formation trajectories.
We also provide a guidance on how to choose an appropriate connection rate in sparse graphs for the trade-off between efficiency and performance. 

Since the spare graph construction entails high computation overhead, we solve the problem offline and current method can only deal with the formation keeping problem with one desired geometric shape.
In furure work, we will speed up this method by pre-generating plenty of sparse graphs as primitives or leveraging learning-based method to enable real-time selection of suitable sparse graphs for varying shapes, and support online formation transformation.
Additionally, we highly desire to know the essential relationships between submatrix selection and formation planning performance in order to enhance our theoretical framework, and plan to integrate our method into actual swarm systems to achieve large-scale formation flight in the wild.

\bibliographystyle{IEEEtran}
\bibliography{bib}

\begin{thebibliography}{10}
\providecommand{\url}[1]{#1}
\csname url@samestyle\endcsname
\providecommand{\newblock}{\relax}
\providecommand{\bibinfo}[2]{#2}
\providecommand{\BIBentrySTDinterwordspacing}{\spaceskip=0pt\relax}
\providecommand{\BIBentryALTinterwordstretchfactor}{4}
\providecommand{\BIBentryALTinterwordspacing}{\spaceskip=\fontdimen2\font plus
\BIBentryALTinterwordstretchfactor\fontdimen3\font minus \fontdimen4\font\relax}
\providecommand{\BIBforeignlanguage}[2]{{%
\expandafter\ifx\csname l@#1\endcsname\relax
\typeout{** WARNING: IEEEtran.bst: No hyphenation pattern has been}%
\typeout{** loaded for the language `#1'. Using the pattern for}%
\typeout{** the default language instead.}%
\else
\language=\csname l@#1\endcsname
\fi
#2}}
\providecommand{\BIBdecl}{\relax}
\BIBdecl

\bibitem{c1}
L.~Quan, L.~Yin, T.~Zhang, M.~Wang, R.~Wang, S.~Zhong, X.~Zhou, Y.~Cao, C.~Xu, and F.~Gao, ``Robust and efficient trajectory planning for formation flight in dense environments,'' \emph{IEEE Transactions on Robotics}, 2023.

\bibitem{c2}
L.~Krick, M.~E. Broucke, and B.~A. Francis, ``Stabilisation of infinitesimally rigid formations of multi-robot networks,'' \emph{International Journal of control}, vol.~82, no.~3, pp. 423--439, 2009.

\bibitem{c3}
B.~D. Anderson, C.~Yu, B.~Fidan, and J.~M. Hendrickx, ``Rigid graph control architectures for autonomous formations,'' \emph{IEEE Control Systems Magazine}, vol.~28, no.~6, pp. 48--63, 2008.

\bibitem{c4}
J.~E. Graver, B.~Servatius, and H.~Servatius, \emph{Combinatorial rigidity}.\hskip 1em plus 0.5em minus 0.4em\relax American Mathematical Soc, 1993, no.~2.

\bibitem{c5}
M.~F. Thorpe and P.~M. Duxbury, \emph{Rigidity theory and applications}.\hskip 1em plus 0.5em minus 0.4em\relax Springer Science \& Business Media, 2006.

\bibitem{c6}
S.~Zhao, Z.~Sun, D.~Zelazo, M.-H. Trinh, and H.-S. Ahn, ``Laman graphs are generically bearing rigid in arbitrary dimensions,'' in \emph{2017 IEEE 56th Annual Conference on Decision and Control (CDC)}, pp. 3356--3361.

\bibitem{c7}
T.~Jord{\'a}n and S.-i. Tanigawa, ``Globally rigid powers of graphs,'' \emph{Journal of Combinatorial Theory, Series B}, vol. 155, pp. 111--140, 2022.

\bibitem{c8}
C.~Boutsidis, M.~W. Mahoney, and P.~Drineas, ``An improved approximation algorithm for the column subset selection problem,'' in \emph{Proceedings of the twentieth annual ACM-SIAM symposium on Discrete algorithms}, 2009, pp. 968--977.

\bibitem{c9}
S.-J. Chung, A.~A. Paranjape, P.~Dames, S.~Shen, and V.~Kumar, ``A survey on aerial swarm robotics,'' \emph{IEEE Transactions on Robotics}, vol.~34, no.~4, pp. 837--855, 2018.

\bibitem{c15}
A.~Kushleyev, D.~Mellinger, C.~Powers, and V.~Kumar, ``Towards a swarm of agile micro quadrotors,'' \emph{Autonomous Robots}, vol.~35, no.~4, pp. 287--300, 2013.

\bibitem{c17}
D.~Morgan, G.~P. Subramanian, S.-J. Chung, and F.~Y. Hadaegh, ``Swarm assignment and trajectory optimization using variable-swarm, distributed auction assignment and sequential convex programming,'' \emph{The International Journal of Robotics Research}, vol.~35, no.~10, pp. 1261--1285, 2016.

\bibitem{c16}
J.~Alonso-Mora, S.~Baker, and D.~Rus, ``Multi-robot formation control and object transport in dynamic environments via constrained optimization,'' \emph{The International Journal of Robotics Research}, vol.~36, no.~9, pp. 1000--1021, 2017.

\bibitem{c18}
X.~Zhou, X.~Wen, Z.~Wang, Y.~Gao, H.~Li, Q.~Wang, T.~Yang, H.~Lu, Y.~Cao, C.~Xu \emph{et~al.}, ``Swarm of micro flying robots in the wild,'' \emph{Science Robotics}, vol.~7, no.~66, p. eabm5954, 2022.

\bibitem{c19}
K.-K. Oh, M.-C. Park, and H.-S. Ahn, ``A survey of multi-agent formation control,'' \emph{Automatica}, vol.~53, pp. 424--440, 2015.

\bibitem{b2}
J.~A. Fax and R.~M. Murray, ``Information flow and cooperative control of vehicle formations,'' \emph{IEEE transactions on automatic control}, vol.~49, no.~9, pp. 1465--1476, 2004.

\bibitem{c21}
S.~Zhao and D.~Zelazo, ``Bearing rigidity and almost global bearing-only formation stabilization,'' \emph{IEEE Transactions on Automatic Control}, vol.~61, no.~5, pp. 1255--1268, 2015.

\bibitem{c23}
Z.~Lin, L.~Wang, Z.~Han, and M.~Fu, ``A graph laplacian approach to coordinate-free formation stabilization for directed networks,'' \emph{IEEE Transactions on Automatic Control}, vol.~61, no.~5, pp. 1269--1280, 2015.

\bibitem{c22}
F.~Xiao, L.~Wang, J.~Chen, and Y.~Gao, ``Finite-time formation control for multi-agent systems,'' \emph{Automatica}, vol.~45, no.~11, pp. 2605--2611, 2009.

\bibitem{c24}
R.~Falconi, L.~Sabattini, C.~Secchi, C.~Fantuzzi, and C.~Melchiorri, ``A graph--based collision--free distributed formation control strategy,'' \emph{IFAC Proceedings Volumes}, vol.~44, no.~1, pp. 6011--6016, 2011.

\bibitem{c25}
C.~C. Cheah, S.~P. Hou, and J.~J.~E. Slotine, ``Region-based shape control for a swarm of robots,'' \emph{Automatica}, vol.~45, no.~10, pp. 2406--2411, 2009.

\bibitem{c27}
Y.~Zhao and P.~A. Vela, ``Good feature matching: Toward accurate, robust vo/vslam with low latency,'' \emph{IEEE Transactions on Robotics}, vol.~36, no.~3, pp. 657--675, 2020.

\bibitem{c26}
M.~Shamaiah, S.~Banerjee, and H.~Vikalo, ``Greedy sensor selection: Leveraging submodularity,'' in \emph{49th IEEE conference on decision and control (CDC)}, 2010, pp. 2572--2577.

\bibitem{c28}
S.~T. Jawaid and S.~L. Smith, ``Submodularity and greedy algorithms in sensor scheduling for linear dynamical systems,'' \emph{Automatica}, vol.~61, pp. 282--288, 2015.

\bibitem{b1}
T.~H. Summers, F.~L. Cortesi, and J.~Lygeros, ``On submodularity and controllability in complex dynamical networks,'' \emph{IEEE Transactions on Control of Network Systems}, vol.~3, no.~1, pp. 91--101, 2015.

\bibitem{c29}
B.~Jackson and T.~Jord{\'a}n, ``Connected rigidity matroids and unique realizations of graphs,'' \emph{Journal of Combinatorial Theory, Series B}, vol.~94, no.~1, pp. 1--29, 2005.

\bibitem{c30}
M.~Tantardini, F.~Ieva, L.~Tajoli, and C.~Piccardi, ``Comparing methods for comparing networks,'' \emph{Scientific reports}, vol.~9, no.~1, p. 17557, 2019.

\bibitem{c33}
N.~B. Priyantha, H.~Balakrishnan, E.~D. Demaine, and S.~Teller, ``Mobile-assisted localization in wireless sensor networks,'' in \emph{Proceedings IEEE 24th Annual Joint Conference of the IEEE Computer and Communications Societies.}, vol.~1, 2022, pp. 172--183.

\bibitem{c31}
J.~H. Holland, ``Genetic algorithms,'' \emph{Scientific american}, vol. 267, no.~1, pp. 66--73, 1992.

\end{thebibliography}

\end{document}